\documentclass[conference]{IEEEtran}
\IEEEoverridecommandlockouts
\usepackage{cite}
\usepackage{amsmath,amssymb,amsfonts}
\usepackage{algorithmic}
\usepackage{graphicx}
\usepackage{textcomp,multirow,multicol}
\usepackage{soul,xcolor}
\def\BibTeX{{\rm B\kern-.05em{\sc i\kern-.025em b}\kern-.08em
    T\kern-.1667em\lower.7ex\hbox{E}\kern-.125emX}}

\usepackage{listings}
\usepackage{tikz}
\usetikzlibrary{arrows}
\usetikzlibrary{calc}
\usetikzlibrary{fit}
\usetikzlibrary{quotes}
\lstdefinelanguage{RBG}{
	alsoletter  = {\#},
    basicstyle=\scriptsize\ttfamily,
    columns=fixed,
    keepspaces=true,
    stepnumber=1,
    numbers=left,
    numbersep=8pt,
    xleftmargin=16pt,
    numberstyle=\color{gray},
    keywords={\#board,\#variables,\#pieces,\#players,\#rules,\#anySquare,\#directedShift,\#queenShift,\#turn},
    keywordstyle=\bfseries\color{purple},
    keywords=[2]{\#anySquare,\#directedShift,\#queenShift,\#turn},
    keywordstyle=[2]\bfseries\color{black},
    morecomment=[l]{//},
    morecomment=[s]{/*}{*/},
    commentstyle=\itshape\color{cyan},
    morestring=[b]",
    stringstyle=\color{teal},
}
\makeatletter
\lstdefinestyle{mystyle}{
  basicstyle=
    \ttfamily
    \small
    \lst@ifdisplaystyle\scriptsize\fi
}
\makeatother
\newtheorem{example}{Example}

\newcommand{\authsp}{}

\begin{document}

\title{Efficient Reasoning in Regular Boardgames
\thanks{This work was supported by the National Science Centre, Poland under project number 2017/25/B/ST6/01920.}
}

\author{
\IEEEauthorblockN{Jakub Kowalski, Rados{\l}aw Miernik, Maksymilian Mika, Wojciech Pawlik,\\ Jakub Sutowicz, Marek Szyku{\l}a, Andrzej Tkaczyk}
\IEEEauthorblockA{\authsp\textit{Institute of Computer Science}\authsp \\
\textit{University of Wroc{\l}aw}\\
Wroc{\l}aw, Poland \\
jko@cs.uni.wroc.pl, radekmie@gmail.com, mika.maksymilian@gmail.com, pawlik.wj@gmail.com,\\ jakubsutowicz@gmail.com, msz@cs.uni.wroc.pl, andrzej.tkaczyk31@gmail.com}
}


\maketitle

\begin{abstract}
We present the technical side of reasoning in Regular Boardgames (RBG) language -- a universal General Game Playing (GGP) formalism for the class of finite deterministic games with perfect information, encoding rules in the form of regular expressions. RBG serves as a research tool that aims to aid in the development of generalized algorithms for knowledge inference, analysis, generation, learning, and playing games. In all these tasks, both generality and efficiency are important.

In the first part, this paper describes optimizations used by the RBG compiler. The impact of these optimizations ranges from 1.7 to even 33-fold efficiency improvement when measuring the number of possible game playouts per second. Then, we perform an in-depth efficiency comparison with three other modern GGP systems (GDL, Ludii, Ai Ai). We also include our own highly optimized game-specific reasoners to provide a point of reference of the maximum speed.
Our experiments show that RBG is currently the fastest among the abstract general game playing languages, and its efficiency can be competitive to common interface-based systems that rely on handcrafted game-specific implementations.
Finally, we discuss some issues and methodology of computing benchmarks like this.
\end{abstract}


\section{Introduction}

The idea of a generalized game playing (GGP) program, the one with the ability to successfully play any given game even such that it has not seen before, may be seen as a direct descendant of the famous General Problem Solver created by Simon, Shaw, and Newell in 1959 \cite{newell1959report}.
Although the first published formalism starting a new domain of GGP research is a work from 1968 by Pitrat \cite{Pitrat1968Realization} concerning a generalization of chess-like games, which was followed in the 90s by Pell and his Metagame approach \cite{Pell1992ANewChallenge}, the real attention towards the idea started in 2005 with the publication of Stanford's Game Description Language (GDL) and the announcement of the annual International General Game Playing Competition (IGGPC) co-located with AAAI conference \cite{Genesereth2005General,Love2008General}.
Since that time, for almost a decade, Stanford's GGP had been the leading field for developing generalized AI solutions, and a source of numerous advancements in search \cite{Finnsson2008Simulation,Finnsson2010Learning}, knowledge representation \cite{Thielscher2010AGeneral,Romero2014Solving}, and other fields \cite{Thielscher2011General,Swiechowski2015Recent}.
In 2016, the last (so far) IGGPC was held, given the number of GDL-related publications was steadily decreasing, as researchers started shifting their attention to other topics.
Today, however, we apparently experience a General Game Playing renaissance.
In just a few years, several alternative languages and platforms had been released -- by multiple author groups, featuring a variety of game types, based on diverse methodologies, and with different purposes under consideration.

These new GGP platforms are made by hobbyists (Ai Ai \cite{TavenerAiAi}), researchers (Regular Boardgames \cite{Kowalski2019RegularBoardgames}, Ludii \cite{piette2020ludii}), or even big companies like Google DeepMind (OpenSpiel \cite{LanctotEtAl2019OpenSpiel}) and Facebook (Polygames \cite{Polygames}).
They range from defining a limited number of boardgames (GBG \cite{Konen2019GBG}), any turn-based games: perfect information deterministic (Regular Boardgames) / nondeterministic with imperfect information (Ludii), to Atari-like real-time games (ALE \cite{Bellemare2013TheArcade}, GVGAI \cite{perez2019general}).
Their methods for describing game rules vary from using regular expressions and automata (Regular Boardgames), a simple objective scripting language (GVGAI), high-level keywords (Ludii), or using underlying game-specific implementations in, e.g., Java (Ai Ai, GBG) / C++ (OpenSpiel, Polygames).
Some are aiming for efficiency, self-containment, and generality under a uniform mechanism (Regular Boardgames), other for human-user game-playing experience (Ai Ai), a study on structure, history, and modeling of games (Ludii), or support for generalized reinforcement learning (OpenSpiel and Polygames).

In this work, we present the technical side of reasoning in Regular Boardgames (RBG) language -- a universal GGP formalism for the class of finite deterministic games with perfect information, encoding rules in the form of regular expressions.
RBG serves as a research tool that aims in development of general algorithms for games, which includes knowledge inference and game analysis, learning, and playing algorithms.
In all these tasks, both generality and efficiency are important.
Generality is necessary to avoid solutions designed only for specific game types, which have no chances to work on a new, and previously unpredicted problem instances.
Computational efficiency makes every task more feasible, allowing e.g., a more detailed analysis of the search tree -- which increases the playing strength of an AI agent.

RBG tries to achieve both goals.
We explain how it reaches a high level of performance, competitive even with some manually implemented reasoners, while still describing games completely in a general abstract form.
We present the insights of the RBG compiler and its optimizations.

Then, we perform an in-depth efficiency comparison with other popular and currently developed GGP systems.
Additionally, we include in the comparison our own highly optimized game-specific reasoners of a few games under RBG interface.
The results from the benchmarks can be used as a point of reference for both implementing a reasoner for a given game and developing new general game playing systems. 
In the former, one can compare the efficiency of a game implementation against various levels of optimization.
As for the latter usage, any GGP approach to be practical requires some amount of fast reasoning. This efficiency survey shows where such a system fits regarding its computational capabilities, and on what types of games it behaves better or worse.
It also points out the set of games that is good to implement when aiming to compare with other GGP systems.
Finally, we discuss issues and methodology of producing such benchmarks, such as the impact of altering the formal game rules for different variants.

\section{Regular Boardgames}

We briefly describe the main concepts of Regular Boardgames. For the full formal definition, we refer to~\cite{Kowalski2019RegularBoardgames}.

A game embedding in RBG consists of a \emph{board}, \emph{variables}, \emph{player roles}, and \emph{rules}.
The \emph{game state} contains a configuration of pieces on the board, values of the variables, the current player, the current position on the board, and the current index (position) in the rules.
The \emph{board} is a directed graph with labeled edges, called \emph{directions}.
The current player, in his turn, can perform a sequence of elementary \emph{actions}, which, when applied sequentially, can modify the game state in a specific way. For an action to be \emph{legal}, it must be both \emph{valid} for the current game state when it is applied and also permitted by the rules.
There are seven types of actions:
\begin{enumerate}
\item \emph{Shift}, e.g., \lstinline|left| or \lstinline|up|, which changes the current position on the board following the specified direction. When there is no such edge, the action is invalid.
\item \emph{On}, e.g., \lstinline|{whiteQueen}|, which does not modify the game state but checks if the specified piece is on the board at the current position.
\item \emph{Off}, e.g., \lstinline|[whiteQueen]|, which puts the specified piece at the current position on the board.
It is always valid.
\item \emph{Comparison}, e.g., \lstinline|{$ turn==100}|, which compares two arithmetic expressions involving variables.
\item \emph{Assignment}, e.g., \lstinline|[$ turn=turn+1]|, which assigns to a variable the value of an arithmetic expression.
\item \emph{Switch}, e.g., \lstinline|->white|, which changes the current player to the specified one. This action ends a move.
\item \emph{Pattern}, e.g., \lstinline|{? left up}|, which is valid only if there exists a legal sequence of actions under the specified rules; in the example, if from the current square there is a path with two edges labeled by \lstinline|left| and \lstinline|up|.
\end{enumerate}

A sequence of actions ending with a switch defines a \emph{move}.
\begin{example}\label{ex:amazons}
In Amazons, the following sequence of actions defines a move with a (white) queen moving two squares up and then shooting an arrow one square right.
{\rm\begin{lstlisting}[numbers=none]
{wQueen} [empty] up up [wQueen] right [arrow] ->black
\end{lstlisting}}
\end{example}
Technically, a \emph{move} is the subsequence of (indexed) actions that are offs, assignments, and switches, together with the positions in rules regular expression where they are applied.
These are precisely the actions that modify the game state, except the board position and the rules index.
Hence, the above example defines a move of length $4$.

A playout ends when the current player has no legal moves. Then, each player's \emph{score} is given in a dedicated variable, named the same as the player's role.
The rules are given by a regular expression over the alphabet of the above actions.
The language defined by this expression contains all potentially legal sequences of actions.
For a concise encoding of the regular expression, a description in RBG is described through C-like macros that are instantiated for given parameters.

\subsection{Example}
A complete example of game Amazons is given in Fig.~\ref{fig:amazons}. Its underlying nondeterministic finite automaton, processed by the game compiler, is shown in Fig.\ref{fig:amazonsnfa}.

At the beginning of the description (Fig.~\ref{fig:amazons}, lines 1--14) we define the players (and their maximal achievable scores), pieces, variables (note that variables for players, containing their current scores, are created automatically), and the board graph with its initial state -- in this case a rectangular board with four possible movement directions.
Then we define some helpful, game-dependent macros. 
\lstinline|anySquare| can change the current position to any square on board, by first jumping an arbitrary number of squares vertically, and then horizontally. 
\lstinline|directedShift| allows movement in direction \lstinline|dir| (given as a parameter) as long as the encountered squares are empty (they contain piece \lstinline|e|), but at least one step has to be made.
\lstinline|queenShift| encodes all possible queen-like moves as a sum of directed shifts. Note that we can pass any sequence of tokens as a macro argument (in this case, two consecutive directions that allow us to encode diagonal movements).

The main game logic, the \lstinline|turn| macro (lines 23--28), encodes a single turn for player \lstinline|me|, whose queen pieces are encoded as \lstinline|piece|. It starts by switching the player to ourselves, then switching the current square to any that contain our queen. We pick up the queen making this square empty, move it to the desired square, and put down. Then we find another square that will be the destination for an arrow. The \lstinline|->>| gives control to the game manager (special role named \emph{keeper}), as the player has no more decisions to make.
The remaining steps put down the arrow (the \lstinline|x| symbol) and set the winning score for the last player. If the playout ends because the current player has no legal moves, this stage will not be reached and the previous player will win the game. 
Finally, the overall rules of the game are encoded as a repetition of the sequence of the white and the black player turns (line~29).

\begin{figure}[ht]
\lstset{numbers=left,columns=flexible,xleftmargin=16pt}
\begin{lstlisting}
#players = white(100), black(100)
#pieces = e, w, b, x
#variables = // no variables
#board = rectangle(up,down,left,right,
         [e, e, e, b, e, e, b, e, e, e]
         [e, e, e, e, e, e, e, e, e, e]
         [e, e, e, e, e, e, e, e, e, e]
         [b, e, e, e, e, e, e, e, e, b]
         [e, e, e, e, e, e, e, e, e, e]
         [e, e, e, e, e, e, e, e, e, e]
         [w, e, e, e, e, e, e, e, e, w]
         [e, e, e, e, e, e, e, e, e, e]
         [e, e, e, e, e, e, e, e, e, e]
         [e, e, e, w, e, e, w, e, e, e])
#anySquare = ((up* + down*)(left* + right*))
#directedShift(dir) = (dir {e} (dir {e})*)
#queenShift = (
    directedShift(up left) + directedShift(up) +
    directedShift(up right) + directedShift(left) +
    directedShift(right) + directedShift(down left) +
    directedShift(down) + directedShift(down right)
  )
#turn(piece; me; opp) = (
    ->me anySquare {piece} [e]
    queenShift [piece]
    queenShift
    ->> [x] [$ me=100, opp=0]
  )
#rules = (turn(w; white; black) turn(b; black; white))*
\end{lstlisting}
\caption{RBG encoding of Amazons (orthodox version, non-splitted).}\label{fig:amazons}
\end{figure}

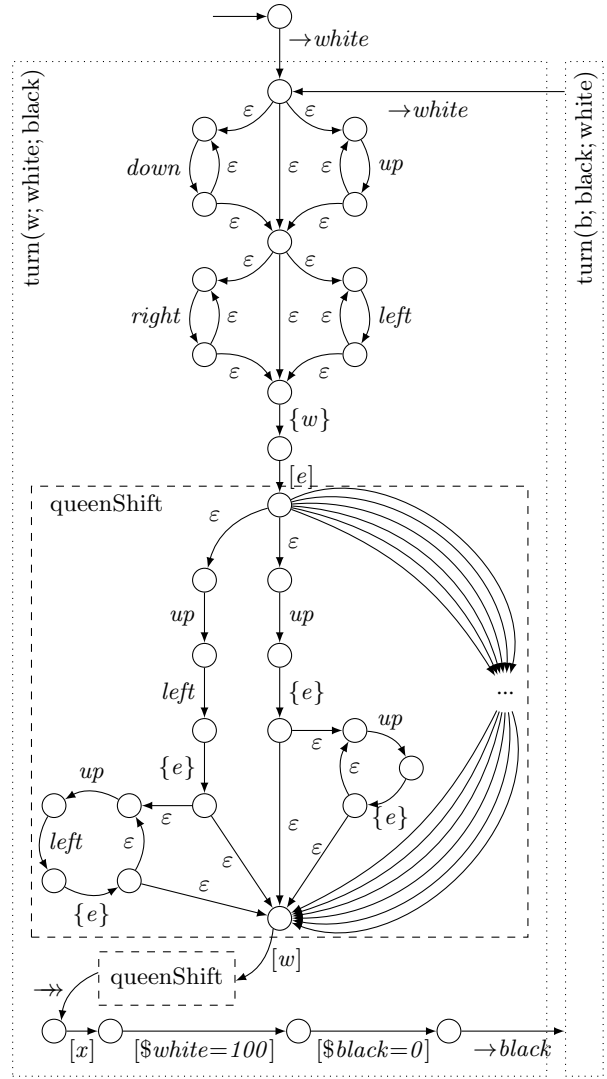
\begin{figure}[ht]\begin{center}
\begin{tikzpicture}\small
\tikzset{edge/.style = {->,> = latex}}
\tikzset{box/.style = {shape=rectangle,minimum size=1em,draw,dashed,fit=#1}}
\tikzset{sphere/.style = {shape=circle,minimum size=1em}}
\tikzset{vertex/.style = {shape=circle,minimum size=1em,draw}}

\node (main00) at (-1,-1) {};
\node[vertex] (main01) at (0,-1) {};
\node[vertex] (main02) at (0,-2) {};
\node[vertex] (main03) at (0,-4) {};
\node[vertex] (main04) at (0,-6) {};
\node[vertex] (main05) at (0,-6.75) {};
\node[vertex] (main07) at (0,-7.5) {};
\node[sphere] (QSdots) at (3,-10) {...};
\node[vertex] (main08) at (0,-13) {};
\node[vertex] (main09) at (-3,-14.5) {};
\node[vertex] (main10) at (-2.25,-14.5) {};
\node[vertex] (main11) at (0.25,-14.5) {};
\node[vertex] (main12) at (2.25,-14.5) {};
\node (main13) at (3.9,-14.5) {};
\node (main14) at (3.9,-2) {};

\node[vertex] (up1) at (1,-2.5) {};
\node[vertex] (up2) at (1,-3.5) {};

\node[vertex] (down1) at (-1,-2.5) {};
\node[vertex] (down2) at (-1,-3.5) {};

\node[vertex] (left1) at (1,-4.5) {};
\node[vertex] (left2) at (1,-5.5) {};

\node[vertex] (right1) at (-1,-4.5) {};
\node[vertex] (right2) at (-1,-5.5) {};

\draw[dashed] (-3.3,-7.25) rectangle (3.3,-13.25);
\node (QStitle) at (-2.3,-7.5) {$\mathrm{queenShift}$};

\coordinate (QSboxA) at (-2.3,-13.55);
\coordinate (QSboxB) at (-0.7,-14.05);
\node[box={(QSboxA) (QSboxB)}] (QSbox) {};
\node at (-1.5,-13.8) {$\mathrm{queenShift}$};

\node[vertex] (QSl1) at (-1,-8.5) {};
\node[vertex] (QSl2) at (-1,-9.5) {};
\node[vertex] (QSl3) at (-1,-10.5) {};
\node[vertex] (QSl4) at (-1,-11.5) {};
\node[vertex] (QSl5) at (-2,-11.5) {};
\node[vertex] (QSl6) at (-3,-11.5) {};
\node[vertex] (QSl7) at (-3,-12.5) {};
\node[vertex] (QSl8) at (-2,-12.5) {};

\node[vertex] (QSu1) at (0,-8.5) {};
\node[vertex] (QSu3) at (0,-9.5) {};
\node[vertex] (QSu4) at (0,-10.5) {};
\node[vertex] (QSu5) at (1,-10.5) {};
\node[vertex] (QSu7) at (1.75,-11) {};
\node[vertex] (QSu8) at (1,-11.5) {};

\draw[dotted] (-3.55,-1.6) rectangle (3.55,-15.1);
\node[rotate=90,anchor=north] (white) at (-3.55,-3.15) {$\mathrm{turn(w; white; black)}$};

\draw[dotted] (3.8,-1.6) rectangle (4.3,-15.1);
\node[rotate=90,anchor=north] (white) at (3.8,-3.15) {$\mathrm{turn(b; black; white)}$};

\draw[edge] (main00) edge (main01);
\draw[edge] (main01) edge node[above right]{$\mathit{{\to}white}$} (main02);
\draw[edge] (main02) edge node[right]{$\mathit{\varepsilon}$} (main03);
\draw[edge] (main03) edge node[right]{$\mathit{\varepsilon}$} (main04);
\draw[edge] (main04) edge node[right]{$\mathit{\{w\}}$} (main05);
\draw[edge] (main05) edge node[right]{$\mathit{[e]}$} (main07);
\draw[edge] (main07) edge[bend left = 19] (QSdots);
\draw[edge] (main07) edge[bend left = 27] (QSdots);
\draw[edge] (main07) edge[bend left = 35] (QSdots);
\draw[edge] (main07) edge[bend left = 44] (QSdots);
\draw[edge] (main07) edge[bend left = 54] (QSdots);zy
\draw[edge] (main07) edge[bend left = 65] (QSdots);
\draw[edge] (QSdots) edge[bend left = 19] (main08);
\draw[edge] (QSdots) edge[bend left = 27] (main08);
\draw[edge] (QSdots) edge[bend left = 35] (main08);
\draw[edge] (QSdots) edge[bend left = 44] (main08);
\draw[edge] (QSdots) edge[bend left = 54] (main08);
\draw[edge] (QSdots) edge[bend left = 65] (main08);
\draw[edge] (main08) edge[bend left] node[right]{$\mathit{[w]}$} (QSbox);
\draw[edge] (QSbox) edge[bend right] node[left]{$\mathit{\to\mathrel{\mkern-13mu}\to}$} (main09);
\draw[edge] (main09) edge node[below]{$\mathit{[x]}$} (main10);
\draw[edge] (main10) edge node[below]{$\mathit{[\$white{=}100]}$} (main11);
\draw[edge] (main11) edge node[below]{$\mathit{[\$black{=}0]}$} (main12);
\draw[edge] (main12) edge node[below]{$\mathit{{\to}black}$} (main13);
\draw[edge] (main14) edge node[below]{$\mathit{{\to}white}$} (main02);

\draw[edge] (main02) edge[bend right] node[above]{$\mathit{\varepsilon}$} (up1);
\draw[edge] (up1) edge[bend left] node[right]{$\mathit{up}$} (up2);
\draw[edge] (up2) edge[bend left] node[left]{$\mathit{\varepsilon}$} (up1);
\draw[edge] (up2) edge[bend right] node[below right]{$\mathit{\varepsilon}$} (main03);

\draw[edge] (main02) edge[bend left] node[above]{$\mathit{\varepsilon}$} (down1);
\draw[edge] (down1) edge[bend right] node[left]{$\mathit{down}$} (down2);
\draw[edge] (down2) edge[bend right] node[right]{$\mathit{\varepsilon}$} (down1);
\draw[edge] (down2) edge[bend left] node[below left]{$\mathit{\varepsilon}$} (main03);

\draw[edge] (main03) edge[bend right] node[above]{$\mathit{\varepsilon}$} (left1);
\draw[edge] (left1) edge[bend left] node[right]{$\mathit{left}$} (left2);
\draw[edge] (left2) edge[bend left] node[left]{$\mathit{\varepsilon}$} (left1);
\draw[edge] (left2) edge[bend right] node[below right]{$\mathit{\varepsilon}$} (main04);

\draw[edge] (main03) edge[bend left] node[above]{$\mathit{\varepsilon}$} (right1);
\draw[edge] (right1) edge[bend right] node[left]{$\mathit{right}$} (right2);
\draw[edge] (right2) edge[bend right] node[right]{$\mathit{\varepsilon}$} (right1);
\draw[edge] (right2) edge[bend left] node[below left]{$\mathit{\varepsilon}$} (main04);

\draw[edge] (main07) edge[bend right] node[above left]{$\mathit{\varepsilon}$} (QSl1);
\draw[edge] (QSl1) edge node[left]{$\mathit{up}$} (QSl2);
\draw[edge] (QSl2) edge node[left]{$\mathit{left}$} (QSl3);
\draw[edge] (QSl3) edge node[left]{$\mathit{\{e\}}$} (QSl4);
\draw[edge] (QSl4) edge node[left]{$\mathit{\varepsilon}$} (main08);
\draw[edge] (QSl4) edge node[below]{$\mathit{\varepsilon}$} (QSl5);
\draw[edge] (QSl5) edge[bend right] node[above]{$\mathit{up}$} (QSl6);
\draw[edge] (QSl6) edge[bend right] node[right]{$\mathit{left}$} (QSl7);
\draw[edge] (QSl7) edge[bend right] node[below]{$\mathit{\{e\}}$} (QSl8);
\draw[edge] (QSl8) edge[bend right] node[left]{$\mathit{\varepsilon}$} (QSl5);
\draw[edge] (QSl8) edge node[above]{$\mathit{\varepsilon}$} (main08);

\draw[edge] (main07) edge node[right]{$\mathit{\varepsilon}$} (QSu1);
\draw[edge] (QSu1) edge node[right]{$\mathit{up}$} (QSu3);
\draw[edge] (QSu3) edge node[right]{$\mathit{\{e\}}$} (QSu4);
\draw[edge] (QSu4) edge node[right]{$\mathit{\varepsilon}$} (main08);
\draw[edge] (QSu4) edge node[below]{$\mathit{\varepsilon}$} (QSu5);
\draw[edge] (QSu5) edge[bend left = 30] node[above]{$\mathit{up}$} (QSu7);
\draw[edge] (QSu7) edge[bend left = 30] node[below]{$\mathit{\{e\}}$} (QSu8);
\draw[edge] (QSu8) edge[bend left] node[right]{$\mathit{\varepsilon}$} (QSu5);
\draw[edge] (QSu8) edge node[above]{$\mathit{\varepsilon}$} (main08);
\end{tikzpicture}
\caption{NFA represented by the Amazons description from Fig.\ref{fig:amazons}.}\label{fig:amazonsnfa}
\end{center}\end{figure}
\section{Optimizations in RBG}\label{sec:optims}

The core of the RBG infrastructure is the compiler, which, given an RBG game description as the input, generates a C++ module implementing a reasoner for this game.
As in every GGP system, the reasoner satisfies a common interface, which, in the case of RBG, allows computing legal moves, reading parameters, accessing the board, etc.

A fundamental part of the reasoner is computing a list of all legal moves. This is done through a DFS-based algorithm \cite[Theorem~9]{Kowalski2019RegularBoardgames} on the automaton that is the NFA representing the game rules joint with the board graph.
A straightforward implementation of the algorithm already provides a decent level of efficiency, but, through a prior analysis of the game rules, we were able to improve it significantly.
We describe here a few of the most important optimizations, which are obtained by inferring knowledge from the game description.
Table~\ref{tab:exp-opt} shows the results of four techniques.

\subsubsection{Shift tables}

Very often, traversing the board consist of multiple shift actions, representing even complex behavior. For instance, on a rectangular board, \lstinline|(left* + right*)(up* + down*)| allows changing the current square into any square, and \lstinline|up* {! up}| changes the square to that in the top row but in the same column.
Obviously, the number of possibilities from such sequences is limited.
Each sequence of actions consisting only of shifts and possibly patterns with shifts can be represented by a map that, for each given square, stores a subset of allowed destination squares.
Hence, we replace each such sequence with one elementary \emph{shift table} action, which simply enumerates all the possibilities with non-deterministic transitions.
Additionally, we can generate further simplifications if the shift table is deterministic or does not depend on the current square.

Shift tables as a whole are the most important optimization, which significantly affects every game.

\subsubsection{Visited check skipping}

The basic reasoning algorithm requires that we check whether a pair of the current square and the index in the rules has been already visited.
Consider for instance \lstinline|((NW + NE + E + SE + SW + W) {x})* {! NW}|, which checks whether from the current square there is a path on squares with \lstinline|x| to the north-west line (example from Hex).
Obviously, by applying actions we could return to the same square and the same position in the rules.
However, in many typical cases, this is not possible.
We can detect these cases by analyzing the transitions in the joint automaton, and omit checking visited pairs.

\subsubsection{Bounding move length}

Because of the \emph{straightness} condition that RBG description must satisfy \cite{Kowalski2019RegularBoardgames}, guaranteeing that the game is finite and the number of legal moves if always finite, the moves have a bounded maximal length measured in the number of modifiers.
For instance, in breakthrough, each move has length 2, which corresponds to the selection of the initial (picking up the pawn) and the destination squares. In chess, the maximum move length is 7.
We can use the limit directly to define the move type structure having exactly this optimal size, also avoiding any dynamic memory allocations.

We can easily calculate this limit if the joint automaton does not contain cycles containing a modifier and not containing a switch.
For such games, we have a usually small upper bound on the length of every move.
In other case (an example is the draughts family of games), such cycles could potentially generate infinite moves, thus the straightness condition must be satisfied non-trivially, and calculating the limit in general is a PSPACE-hard problem (cf.\ \cite[Theorem~10]{Kowalski2019RegularBoardgames}).

\subsubsection{Monotonic classes of moves}

Sometimes, especially in simple games, the bottleneck is the general interface itself. In the case of RBG compiler, all legal moves must be generated every time from scratch. Besides advantages like preserving minimal game state representation, informative move content (which contains a sequence of actions, in contrast to, e.g., a pure move index), fixed ordering, and modifiable moves list, it comes with an efficiency drawback in certain situations.
In the case of simple games with many moves, the cost of generating them can be prevailing and thus get behind systems that admit, e.g., only modifying the list of legal moves.

We develop the general concept of \emph{monotonicity classes}, which can deal with the above problem in several game types.
We will split the game states into classes such that they share their legal moves.
Let $\mathbb{S}$ be the set of all reachable game states from the initial state, and for $S \in \mathbb{S}$, let $\mathrm{moves}(S)$ be the set of all legal moves.
Let $\mathbb{M} = \bigcup_{S \in \mathbb{S}} \mathrm{moves}(S)$ (the set of all possible moves).
Now we define that a function $c\colon \mathbb{S} \to \mathbb{N}$ is \emph{monotonic classifier} if for every state $S \in \mathbb{S}$, we have $\mathrm{moves}(S') \subseteq \mathrm{moves}(S)$ for every state $S' \in \mathbb{S}$ that is a successor of $S$ in the game tree with $c(S)=c(S')$.
There always exists a trivial monotonic classifier that assigns a different class number to each state.
However, the smaller number of assigned classes is better.
A game description is $k$-move-monotonic if there exists a monotonic classifier assigning at most $k$ class numbers.
Obviously, the existence of monotonic classifiers depends on the particular move representation.

Returning to RBG, a natural candidate to classify moves are switches.
For each switch, we need to check whether for a game state with the rules index at this switch, the legal moves are a superset of the legal moves of every successor game state.
We use several conditions for that, for example, if a move is related to the specific content of a square (e.g., empty) and, in the rules, this content is never added, then there will be no new moves in the successor game states.
The monotonicity optimization also requires shift tables to detect if moves do not depend on the current square.
Using that, we can determine that the game descriptions of, e.g., Connect4, Gomoku, and Hex (without the pie rule) are 2-monotonic.
Also, in Pentago (split), we assign one monotonic class for the moves related to rotation (the eight rotation moves are invariant), despite that the placement moves cannot be assigned to the same class.

\subsection{Efficiency Gain}

\begin{table*}[!htb]\renewcommand{\arraystretch}{1.05}\small
\newcommand{\col}[1]{\multicolumn{1}{c|}{#1}}
\newcommand{\colt}[1]{\multicolumn{2}{c|}{#1}}
\newcommand{\colttI}[1]{\multicolumn{3}{c||}{#1}}
\newcommand{\coltttt}[1]{\multicolumn{5}{c|}{#1}}
\newcommand{\colr}[1]{\multicolumn{2}{r|}{#1}}
\newcommand{\colrr}[1]{\multicolumn{3}{r|}{#1}}
\newcommand{\tb}[1]{\textbf{#1}}
\newcommand{\nb}[1]{#1}
\caption{The impact of specific optimizations of the RBG compiler (Flat MC playouts/sec.).}\label{tab:exp-opt}
\begin{center}\begin{tabular}{|l||r|r|r|r|r|r|r|r|}\hline
\multirow{2}{*}{\bf Game}&\col{\bf No }&\col{\bf No shift tables,}&\col{\bf No visited}&\col{\bf No bounding}&\col{\bf No monotonic}&\col{\bf All}\\
                         &\col{\bf optimizations}&\col{\bf no mon. classes}&\col{\bf check skipping}&\col{\bf move length}&\col{\bf classes}&\col{\bf opt.}\\\hline
Amazons             &   1,642 (-41\%) &   2,500 (-10\%) &     2,144 (-23\%) &     2,078 (-25\%) &           (  0\%) &   2,781    \\\hline
Amazons (split2)    &   9,340 (-48\%) &  12,264 (-32\%) &    14,818 (-18\%) &    15,682 (-13\%) &           (  0\%) &  18,084   \\\hline
Arimaa (split)      &      79 (-91\%) &     112 (-88\%) &       751 (-16\%) &           (  0\%) &           (  0\%) &     898   \\\hline
Breakthrough (8x8)  &  16,330 (-63\%) &  21,022 (-52\%) &    29,136 (-33\%) &    40,269 ( -8\%) &           (  0\%) &  43,575   \\\hline
Canadian Draughts   &     442 (-70\%) &     449 (-69\%) &     1,294 (-12\%) &           (  0\%) &           (  0\%) &   1,465   \\\hline
Chess (50-move rule)&     249 (-73\%) &     370 (-60\%) &       656 (-30\%) &       854 ( -9\%) &           (  0\%) &     935    \\\hline
Connect4            & 271,240 (-66\%) & 351,767 (-56\%) &   604,614 (-25\%) &   765,700 ( -5\%) &   485,451 (-40\%) & 804,326  \\\hline
English Draughts    &  14,052 (-75\%) &  14,327 (-75\%) &    30,593 (-46\%) &           (  0\%) &           (  0\%) &  56,269   \\\hline
Gomoku (standard)   &   3,455 (-97\%) &   5,377 (-95\%) &    81,801 (-28\%) &    95,561 (-16\%) &     7,101 (-94\%) & 113,718  \\\hline
Knightthrough       &  27,193 (-59\%) &  35,254 (-46\%) &    47,637 (-28\%) &    52,469 (-21\%) &           (  0\%) &  65,823   \\\hline
Pentago (split)     &  16,854 (-63\%) &  20,782 (-54\%) &    43,207 ( -5\%) &  44,993 ( -1\%) &    42,942 ( -6\%) &  45,445   \\\hline
Tic-tac-toe         & 767,315 (-57\%) & 962,030 (-46\%) & 1,550,360 (-13\%) & 1,575,951 (-11\%) & 1,374,291 (-23\%) &  1,777,036   \\\hline
\end{tabular}\end{center}
\end{table*}

Table~\ref{tab:exp-opt} shows the impact of the described optimizations.
We show their importance in the final version by the efficiency drop if an optimization is skipped.
The effects of the optimizations strongly correlate.
Monotonic classes optimization requires shift tables, thus there is no result for a variant with only shift tables skipped. 
Also, as described before, some optimizations (monotonic classes, bounding move length) provide a boost only for specific types of games, and are neutral for all the remaining ones; these cases are represented by $0\%$.

The most significant and universal optimization is shift tables.
The second one, not much behind it, is visited checks optimization.
They both affect every game.
Bounding move length is a decent optimization and affects only games where moves have a bounded length by the rules and independently on the board, but this is actually a large class of games.
Monotonic classes affect only simple games where moves are very straightforward, but these are usually the cases where listing all moves adds a significant computational cost.

Optimizations positively affect the computation time because they mostly reduce the amount of the generated C++ code, whose compilation is by far dominant.
For example, Chess with optimizations is compiled in 7.2s and without them in 10.43s.
In general, all the first three optimizations reduce the compilation time, and monotonic classes leave it unaffected.
Except for monotonic classes, which require to store moves along with the game state, the optimizations do not have any real drawbacks, thus they should always be used.

\section{Comparison of Different Systems}

\subsection{Other GGP Approaches}

Here we present in slightly more detail GGP systems that will be used in our experiments.
We can describe those systems as belonging to three types of GGP approaches: ``true'' general game playing, where the description language is ``closed'' (e.g., GDL, Toss \cite{Kaiser2011FirstOrder}, Regular Boardgames); ``hybrid'', that describe games using an extendable set of generalized keywords (Metagame \cite{Pell1992METAGAME}, Ludi, VGDL \cite{Schaul13}, Ludii); and one that just make use of a common interface for game-specific implementations (Ai Ai, GBG, OpenSpiel, Polygames).
These categories are informal. Closed languages try to provide a uniform and minimal mechanism so that each new game can be
effectively implemented purely in the proposed language.
Hybrid languages try to provide high-level concepts that cover parts of game rules. As such, implementing a new game that requires a new rule type usually needs an extension of the language.
The last type of systems just requires games to be manually implemented in a usual programming language and satisfying some interface.
They also often provide parameterization of the rules.
In theory, more game-specific code allows more optimization, thus should result in higher efficiency.

The GGP systems we have chosen for the comparison are the ones that are possibly very recent, currently developed, and containing enough games to find a common test set, with the exception of GDL, which is a classical example. Besides Regular Boardgames, there has been no recent approach to create a closed language for describing a large class of games. 

We also performed experiments with OpenSpiel \cite{LanctotEtAl2019OpenSpiel}, however, given that this system during the playouts also computes observation tensors makes the comparison unfair. Thus, we decided not to include the results in this paper.

Apart from the other GGP systems that we described above, for some games we had developed game-specific reasoners (in C++) that implement the common RBG interface (currently, the part of it necessary for computing moves and states).
This is an attempt to show possibly maximal reasoning efficiency.
The implementations are highly optimized with a lot of low-level tricks designed for a single specific game.

\begin{table*}[!htb]\renewcommand{\arraystretch}{1.05}\small
\newcommand{\colt}[1]{\multicolumn{2}{c|}{#1}}
\newcommand{\colttI}[1]{\multicolumn{3}{c||}{#1}}
\newcommand{\coltttt}[1]{\multicolumn{5}{c|}{#1}}
\newcommand{\tb}[1]{\textbf{#1}}
\newcommand{\nb}[1]{#1}
\newcommand{\n}{\hphantom{*}}
\newcommand{\fb}{\textdagger}
\newcommand{\m}{\hphantom{\textdagger}}
\caption{Comparison of the reasoning efficiency of different GGP systems. The percentage values are ratios to the RBG compiler \hfill \newline (Flat MC playouts/sec.).}\label{tab:exp-comparison}
\begin{center}\begin{tabular}{|l||r|r|r|r|r|r|r|}\hline
{\bf Game}&{\bf GDL propnet}      &{\bf Ludii 0.9.3}       &{\bf Ai Ai 4.0.2.0}          &{\bf RBG compiler 1.2}         &{\bf RBG game-specific 1.2} \\\hline
Amazons                    &      4 (0.1\%) &            --\n &                --\n &        2,781\m & --\m \\\hline
Amazons (split2)           &    365  ( 2\%) &  2,634 (15\%)\n &    13,724 ( 76\%)\n &       18,084\m & --\m \\\hline
Arimaa (split)             &             -- &     22 ( 2\%)*  &      4,507 (501\%)* &          898\m & --\m \\\hline
Breakthrough (8x8)         &  2,711  ( 6\%) &   2,344 ( 5\%)\n&    29,247 ( 67\%)\n &       43,575\m & 157,333 (361\%)\m\\\hline
Canadian Draughts          &             -- &    156 (11\%)*  &                --\n &        1,465\m & --\m \\\hline
Chess (50-move rule)       &     43  ( 5\%) &     88 ( 9\%)*  &    248     ( 27\%)* &          935\m & --\m \\\hline
Connect4                   & 46,896  ( 6\%) & 38,544 ( 5\%)\n & 1,315,457 (164\%)\fb&      804,326\fb& 2,139,403 (266\%)\fb\\\hline
Connect6 (split)           &             -- &  1,192 ( 3\%)\n &    21,725 ( 55\%)\n &       39,330\m & --\m \\\hline
English Draughts           &             -- &            --\n &                --\n &       56,269\m & 188,143 (334\%)\m\\\hline
English Draughts (split)   &  3,429  ( 6\%) &   2,830 (5\%)*  &    84,751 (143\%)*  &       59,335\m & 231,252 (390\%)\m\\\hline
Gomoku (standard)          &  1,147  ( 1\%) &  4,091 ( 4\%)\n &    47,332 ( 42\%)\n &      113,718\m & --\m \\\hline
Hex (9x9)                  &    476 (0.8\%) &  9,259 (16\%)\n &    95,113 (165\%)\n &       57,508\m & --\m \\\hline
Knightthrough (8x8)        &  4,913  ( 7\%) &  2,987 ( 5\%)\n &    68,250 (104\%)\n &       65,822\m & --\m \\\hline
Pentago (split)            &  6,408  (14\%) &            --\n &                --\n &       45,445\m & --\m \\\hline
Reversi                    &     370  (3\%) &    757 ( 5\%)*  &    53,866 (387\%)*  &       13,910\m & 182,228 (1,310\%)\m\\\hline
Skirmish (100 turns)       &     239  (3\%) &    848 (11\%)*  &                --\n &        7,715\m & --\m \\\hline
Yavalath                   &             -- & 49,060 ( 8\%)\n &   204,484 ( 32\%)\n &      636,032\m & --\m \\\hline
\end{tabular}\end{center}\begin{flushleft}
*\ --\ the implemented rules are different from the version in RBG (explained in Subsection~\ref{subsec:rules}).\\
\fb\ --\ see the issues described in Subsection~\ref{subsec:randgen}.
\end{flushleft}\end{table*}

\subsubsection{Stanford's GDL}
GDL~\cite{Love2008General}, used in IGGPC, is the most well-known and deeply-researched game description language. It can describe any turn-based, simultaneous-moves, finite, and deterministic $n$-player game with perfect information. It is a high-level, strictly declarative language based on Datalog.

GDL does not provide any predefined functions, meaning that every predicate encoding the game structure must be defined explicitly from scratch.
As a result, the game descriptions are usually long and hard to understand. Because their processing requires logic resolution, it is also very computationally expensive.
In fact, many games expressible in GDL could not be played by any program at a decent level. Some games, due to computational cost, are not playable at all.
For example, features like longest ride in checkers or capturing in go are difficult and inefficient to implement.
In such cases, only simplified rules are encoded (yet often they are available in repositories under the standard name of the full game).
GDL has a number of independent reasoner implementations, among which propnets \cite{Sironi2016Optimizing} are considered the fastest.

\subsubsection{Ludii}
Although the Ludii system \cite{piette2020ludii} (the successor of Ludi, used to generate first market-selling AI-authored boardgame \cite{Browne2010Evolutionary}) has been designed primarily to chart the historical development of games and explore their role in human culture, its latest versions came out with additional tools for agent implementations, game visualizations and human playing \cite{stephenson2019ludiiCompetition}.
The language is based on a large number of ludemes, conceptual units of
game-related information, whose behavior is encoded in the underlying Java implementation.
This makes the resulting games usually concise and well suited for tasks such as procedural content generation, but hard to understand without documentation of each ludeme, which is already very long and constantly growing.
Another drawback is that a large but limited set of currently implemented ludemes greatly hampers natural expressiveness and efficiency of more complex and non-standard games that do not have dedicated highly specialized building-blocks.
On the positive sides, Ludii comes with a large number of implemented games.
The language allows generation of generalized game-related content such as human-playable GUI and various game/algorithm analyzing tools.
High-level ludemes are also an easy source of heuristics, which Ludii agents can benefit, without the need to detect game features in a knowledge-free manner.

The efficiency is similar to that of a GDL propnet, sometimes overcoming the latter.
Ludii is closed-source with one reference implementation provided, and due to the complication level, it is impractical to develop an independent branch.

\subsubsection{Ai Ai}
Stephen Tavener's Ai Ai \cite{TavenerAiAi} is a closed source program that allows playing abstract games versus both AI and human opponents, with user-friendly visualization, multiple options to customize, AI settings, and game analysis tools. It is an advanced platform containing many games, and more are being added all the time. 
Games can be hand-coded in Java (for efficiency), or assembled from large blocks using the MGL (Modular Game Language) -- a scripting language based on JSON.
In practice, almost all of the games are programmed directly in Java, so the resulting game engine is as fast as its underlying implementation is optimized. Thus, although it is considered as a general game playing approach, the reasoners are game-specific with a common interface.

\subsection{Technical Setup}

All experiments were performed on a single core of Intel(R) Core(TM) i7-4790 @3.60GHz of a computer with 16GB RAM.
The GCC version was \texttt{gcc (Ubuntu 7.5.0-3ubuntu1~18.04) 7.5.0} with \texttt{boost 1.65.1.0}.
The Java version was \texttt{Java(TM) SE Runtime Environment (build 13.0.2+8)}.

Each test (one game) was a run of the flat MC algorithm, yielding statistics of the average score of uniformly random playouts for each legal move from the initial state of the game, or just a run of random playouts for a given time, depending on the system.
The preprocessing time was not counted in any case.
The GGP system versions were the available ones up-to-date on 4th June, 2020.

Each GDL propnet test constitutes of the average time of 10 runs lasting 10 minutes, not counting the preprocessing (averaging is a proper practice here, because of non-deterministic propnet construction).
The used propnet implementation is by C.~Sironi based on \texttt{ggp-base}, currently not available online, but some results were reported independently \cite{Sironi2016Optimizing}.

Each Ludii~0.9.3 test was performed via the command-line option \texttt{{-}{-}time-playouts} with default settings and 1~minute measuring time.

Each Ai Ai~4.0.2.0 test was performed through the dedicated menu option \texttt{MC Playouts/Second (This Game)}, which measures over 100 seconds.

Each RBG~1.2 test lasted 1~minute and was performed via the shared benchmark script (\texttt{ rbg2cpp/run\_benchmark.sh}).
A test of an RBG game-specific reasoner was also 1~minute long, and it was compiled with the same overlaying benchmark procedure used for RBG; the package is included in RBG~1.2 release.
This version contains all optimizations described in Section~\ref{sec:optims}.

\subsection{The Results}

Table~\ref{tab:exp-comparison} shows the results of the main experiment.
There is a large gap between systems with abstract languages (GDL, Ludii) and systems with game-specific implementations (Ai Ai).
RBG achieves similar performance to the latter, although the values strongly vary depending on the game, which could be explained by various levels of effort put in optimizing a game-specific implementation.
Our game-specific implementations are faster than almost everything else, showing that automatically generated RBG reasoners still have optimization potential, as the RBG interface is currently not a barrier.

\section{Impact of Methodology}

In \cite{Kowalski2020Experimental} we pointed out several issues concerning the methodology of the benchmarks in GGP; here, we discuss two technical ones that particularly affected our experiment.

\subsection{Influence of the Rules Implementation}\label{subsec:rules}

An important issue is game matching, which needs special care among different systems.
By the \emph{same games} we understand those that have isomorphic game trees.
This includes win/draw/loss distinction in terminal states.
We made an effort to match the games in RBG with the existing implementations in GDL.
In the other systems, some games have encoded a different variation of the rules. Up to our knowledge, we marked all these cases in Table~\ref{tab:exp-comparison} with a star.
These differences are relatively minor to provide a meaningful comparison, basing on our subjective opinion and some internal tests with game variations.
A possible exception is Chess and Arimaa in Ai Ai; they implement, among others, the threefold repetition rule, which is costly. 
Also, Canadian Draughts in Ludii is a split version. 
Nevertheless, the results could differ slightly under an exact match.
Example: English Draughts (split) in RBG and GDL ends in a draw after 20 moves without moving a man nor a capture.
However, in Ludii, instead of that, there is an internal hard turn limit set independently on the rules.
This is a minor difference, as ending a random playout in this way is rare.
As it is sadly not a standard for GGP systems to provide exact specification of the rules or reliable game statistics, most of such disparities are very hard to spot, thus they influence the fairness of the published benchmarks. 
Here, we would like to show some more detailed examples of how heavily the reasoning efficiency can be altered by modifying the game rules without changing their commonsense meaning.

Let us continue our example of Amazons. The orthodox version under the standard interpretation is that the player's single turn consists of moving a queen and shooting an arrow. Thus, the first player has 2176 possible moves, and the average branching factor is 374 for the first player and 299 for the second \cite{hensgens2001knowledge}. 
However, some implementations modify the rules so the player turn is split in two: firstly a queen movement is selected, and then an arrow shot from this queen. 
This interpretation operates on the game tree that is not isomorphic with the orthodox version, but it considerably reduces the branching factor thus computation time \cite{kloetzer2007monte}.
The rules in RBG encoding the described variant are shown in Fig.~\ref{fig:amazonsvariants}.
Compared to the orthodox version, in RBG, this \emph{split2} approach allows more than 6 times faster simulations (see Table~\ref{tab:amazons-comparison}).

Although \emph{split2} is, thanks to its straightforwardness, the most popular unorthodox variant, there are many other possible reinterpretations of the rules. Another example based on splitting player's move into two parts (\emph{split2a} in Tab.~\ref{tab:amazons-comparison}) chooses a queen and its movement direction in its first part, and shifts all the remaining operations to the second part. This version reaches similar efficiency as its predecessor.
However, it is possible to create other variants that will be significantly faster. 
A variant named \emph{split5} (see Fig.~\ref{fig:amazonsvariants} for its rules) starts a new turn after nearly every atomic choice that guarantees the correctness of the remaining playout.
This variant is over two times faster than \emph{split2}.
Table~\ref{tab:amazons-comparison} shows the results for even more variants based on the same orthodox encoding of Amazons, visualizing the possible impact of the split-based trick we described. All mentioned Amazon variants are available in the RBG repository; they are obtained by a minor modification of the encoding of the orthodox version.

\begin{figure}[htb]\small
\hspace{1em} {\footnotesize The split2 variant of amazons (difference code):}
\lstset{numbers=left,firstnumber=23,xleftmargin=16pt}
\begin{lstlisting}
#turn(piece; me; opp) = (
    ->me anySquare {piece} [e]
    queenShift [piece]
    ->me queenShift
    ->> [x] [$ me=100, opp=0]
  )
\end{lstlisting}
\hspace{1em} {\footnotesize The split5 variant of amazons (difference code):}
\lstset{numbers=left,xleftmargin=16pt,firstnumber=16}
\begin{lstlisting}[belowskip=-0.8 \baselineskip]
#directedShift(dir; me) = (dir {e} ->me (dir {e})*)
\end{lstlisting}
\lstset{numbers=left,xleftmargin=16pt,firstnumber=23}
\begin{lstlisting}
#turn(piece; me; opp) = (
    ->me anySquare {piece} {? anyNeighbor {e}} ->> [e]
    ->me queenShift(me) ->> [piece]
    ->me queenShift(me) ->> [x] [$ me=100, opp=0]
  )
\end{lstlisting}

\caption{Two unorthodox variants of Amazons in RBG.}\label{fig:amazonsvariants}
\end{figure}

\begin{table}[htb]\renewcommand{\arraystretch}{1.05}\small
\newcommand{\colt}[1]{\multicolumn{2}{c|}{#1}}
\newcommand{\colttI}[1]{\multicolumn{3}{c||}{#1}}
\newcommand{\coltttt}[1]{\multicolumn{5}{c|}{#1}}
\newcommand{\tb}[1]{\textbf{#1}}
\newcommand{\nb}[1]{#1}
\caption{Comparison of efficiency of different variants of Amazons (Flat MC playouts/sec.).
}\label{tab:amazons-comparison}
\begin{center}\begin{tabular}{|l||r|r|r|r|}\hline
{\bf Game}&{\bf RBG 1.2} &{\bf speedup}\\\hline
Amazons (orthodox)&  2,781 &  100\% \\\hline
Amazons (split2)  & 18,084 &  650\% \\\hline
Amazons (split2a) & 18,108 &  651\% \\\hline
Amazons (split3)  & 34,934 &  1,256\% \\\hline
Amazons (split5)  & 38,694 &  1,391\% \\\hline
Amazons (split5+) & 38,004 &  1,367\% \\\hline  
\end{tabular}
\end{center}
\end{table}

\subsection{Random Generators and Benchmark Procedures}\label{subsec:randgen}

\begin{table*}[!htb]\renewcommand{\arraystretch}{1.05}\small
\newcommand{\colt}[1]{\multicolumn{2}{c|}{#1}}
\newcommand{\coltt}[1]{\multicolumn{3}{c|}{#1}}
\newcommand{\colttI}[1]{\multicolumn{3}{c||}{#1}}
\newcommand{\coltttt}[1]{\multicolumn{5}{c|}{#1}}
\newcommand{\tb}[1]{\textbf{#1}}
\newcommand{\nb}[1]{#1}
\caption{The impact of the used random generator (Flat MC playouts/sec.).}\label{tab:generators}
\begin{center}\begin{tabular}{|l||r|r|r|r|r|r|r|}\hline
\multirow{2}{*}{\bf Game}&\coltt{\bf RBG compiler}&\coltt{\bf RBG game-specific} \\\cline{2-7}
                        & Default method & Java method & Lemire's method & Default method & Java method & Lemire's method \\\hline
Breakthrough            &         43,575 &      42,738 &          34,917 &        157,332 &     182,547 &         144,175 \\\hline
Connect4                &        804,326 &   1,052,897 &       1,075,988 &      2,139,403 &   4,230,855 &       4,965,320 \\\hline
English Draughts        &         56,269 &      58,615 &          59,044 &        188,143 &     249,169 &         251,900 \\\hline
English Draughts (split)&         59,335 &      62,506 &          65,182 &        231,252 &     295,987 &         296,895 \\\hline
Reversi                 &         13,910 &      13,961 &          14,140 &        182,228 &     213,313 &         219,012 \\\hline
\end{tabular}
\end{center}
\end{table*}

When doing many simulations, the overlaying interface becomes meaningful. We show this on a particular part that is the random generator used to draw moves uniformly in flat MC.
This issue has never been raised before, but it is quite noticeable when the number of turns per second is large enough.
In Table~\ref{tab:generators}, we demonstrate the possible impact of the generator, which includes the random generator itself and an unbiased method for drawing an integer from a range.
There is the standard method combining \texttt{std::uniform\_int\_distribution} with \texttt{std::mt19937} (used in the tests for Tables~\ref{tab:exp-opt}--\ref{tab:amazons-comparison}), a reimplemented Java method from \texttt{java.util.Random}, and a modern unbiased drawing algorithm by Lemire \cite{Lemire2018FastRandom} combined with a fast Mersenne Twister \texttt{boost::random::mt11213b}.

In our experiments, for RBG, we have used the default method, which is usually the slowest of the three but probably of the highest quality (based on the traditional measurement of the period).
The propnet, Ai Ai, and most likely also Ludii, use the standard Java generator.
Of course, there is a trade-off between the quality and the speed, and different systems use different methods.
From our experience, the choice of reasonable generator does not influence the quality of agent nor change the statistics, but it impacts the cost of computing.
The impact becomes higher when the reasoning itself is faster.
In extreme cases, as Connect4, the cost of random move selection can be dominating.

The issue does not concern only random generators, but the whole benchmark procedure (with time measurements, gathering statistics, etc.).
For instance, the flat MC algorithm in Ai Ai for Connect4 performs a much larger number of iterations per second (we got even 3,428,427) than the benchmark report, while other, more costly games reveal no noticeable difference.

Concluding, when reaching such a performance level, it is difficult to provide a reliable benchmark.
Nevertheless, in such cases, we can expect that the cost of reasoning would be negligible compared to any accompanying computation, and then, the efficiency of reasoning loses its importance.

\section{Conclusion}

Regular Boardgames is a modern general game playing system aiming for efficiency and describing games via an abstract, concise, and well-defined formal language.
The shared environment currently consists, in particular, of the game compiler to C++, a network-based game manager, and a high-level API allowing writing AI in Python.
In this paper, we have described a few optimizations of the RBG compiler, as one of the sources of its efficiency.

We performed extensive experiments comparing the efficiency of five modern general game playing systems.
We conclude that RBG significantly outperforms systems based on other abstract languages and has comparable (with a high variation) performance to game-specific reasoners of other systems as Ai Ai. 
By comparing with our hand-made game-specific reasoners under the same interface, we demonstrated that there is still potential for optimization.
This leads to the following research question: given game rules, how to automatically produce an optimal reasoner?
Our implemented optimizations so far are just an infantile play around it.

The final issues discussed, so far overlooked, should help in developing standardized benchmark methods concerning reasoners, which would allow fair, reproducible, and transparent comparisons.


\section*{Acknowledgments}

We thank Stephen Tavener for helping and explaining how to perform the benchmark of his Ai Ai system.
We also thank Chiara F.\ Sironi for sharing the GDL propnet code and for helping with using it. 
Finally, we thank anonymous reviewers for their valuable comments and important insights.

\bibliographystyle{IEEEtran}
\bibliography{bibliography}

\end{document}